\pdfoutput=1

\documentclass[11pt]{article}

\usepackage[]{acl}

\usepackage{times}
\usepackage{latexsym}
\usepackage{dsfont}
\usepackage{graphicx}
\usepackage{anyfontsize}
\usepackage{booktabs} 
\usepackage{multirow}
\usepackage{url}
\usepackage{amsmath}
\usepackage{amsfonts}
\usepackage{graphicx}
\usepackage{bm}
\usepackage{color}
\usepackage{diagbox}
\usepackage{xspace}
\usepackage{subfigure}

\usepackage[T1]{fontenc}

\usepackage[utf8]{inputenc}

\usepackage{microtype}

\title{Prompt Tuning for Discriminative Pre-trained Language Models}

\author{Yuan Yao$^{1}$, Bowen Dong$^{1}$, Ao Zhang$^{2}$, Zhengyan Zhang$^{1}$, \\
\textbf{Ruobing Xie$^{3}$, Zhiyuan Liu$^{1,4\dagger}$, Leyu Lin$^{3}$, Maosong Sun$^{1,4\dagger}$, Jianyong Wang$^{1}$}\\
        $^1$Dept. of Comp. Sci. \& Tech., Institute for AI, Tsinghua University, Beijing, China
        \\
         Beijing National Research Center for Information Science and Technology\\
         $^2$Department of Computer Science, National University of Singapore, Singapore \\ 
         $^3$WeChat Search Application Department, Tencent, China \\ 
         $^4$Institute for Artificial Intelligence, Tsinghua University, Beijing, China \\
         Institute Guo Qiang, Tsinghua University, Beijing, China \\
         International Innovation Center of Tsinghua University, Shanghai, China \\
         \texttt{yaoyuanthu@163.com} \ \ \  \texttt{dongbw18@mails.tsinghua.edu.cn}\\
}

\begin{document}
\maketitle
\begin{abstract}
Recent works have shown promising results of prompt tuning in stimulating pre-trained language models (PLMs) for natural language processing (NLP) tasks. However, to the best of our knowledge, existing works focus on prompt-tuning generative PLMs that are pre-trained to generate target tokens, such as BERT~\cite{DBLP:conf/naacl/DevlinCLT19}. It is still unknown whether~and how discriminative PLMs, e.g., ELECTRA~\cite{DBLP:conf/iclr/ClarkLLM20}, can be effectively prompt-tuned. In this work, we present DPT, the first \underline{p}rompt \underline{t}uning framework for \underline{d}iscriminative PLMs, which reformulates NLP tasks into a discriminative language modeling problem. Comprehensive experiments on text classification and question answering show that, compared with vanilla fine-tuning, DPT achieves significantly higher performance, and also prevents the unstable problem in tuning large PLMs in both full-set and low-resource settings. The source code and experiment details of this paper can be obtained from \url{https://github.com/thunlp/DPT}.
\end{abstract}

{\let\thefootnote\relax\footnotetext{$^\dagger$ Corresponding authors: Z.Liu (liuzy@tsinghua.edu.cn), M.Sun (sms@tsinghua.edu.cn)}}

\section{Introduction}
Recent years have witnessed the great success of the \textit{pre-training-then-fine-tuning} paradigm in natural language processing (NLP)~\cite{DBLP:conf/naacl/DevlinCLT19,DBLP:conf/nips/YangDYCSL19,DBLP:conf/iclr/ClarkLLM20,DBLP:conf/iclr/LanCGGSS20,DBLP:journals/jmlr/RaffelSRLNMZLL20}. Typically, language models are first pre-trained on large-scale corpora via self-supervised generative or discriminative tasks to learn universal text representations, and then fine-tuned to adapt to downstream tasks~\cite{qiu2020pre,xu2021pre}. However, the significant gap between the objective forms of model pre-training and fine-tuning hinders taking full advantage of PLMs in downstream tasks~\cite{DBLP:journals/corr/abs-2107-13586}.

Prompt tuning has recently shown its effectiveness in stimulating the capability of PLMs by transforming downstream tasks into the same form as pre-training~\cite{DBLP:conf/emnlp/PetroniRRLBWM19,brown2020language,DBLP:conf/naacl/SchickS21,DBLP:conf/acl/GaoFC20,yao2021cpt}. However, to the best of our knowledge, existing works focus on prompt-tuning generative PLMs (i.e., PLMs pre-trained by generating target textual tokens from the context, such as BERT~\cite{DBLP:conf/naacl/DevlinCLT19} and GPT~\cite{brown2020language}). It is still unknown whether and how discriminative PLMs can be effectively prompt-tuned (i.e., PLMs pre-trained by discriminating replaced tokens, such as ELECTRA~\cite{DBLP:conf/iclr/ClarkLLM20} and WKLM~\cite{DBLP:conf/iclr/XiongDWS20}). Since discriminative PLMs typically enjoy competitive performance and superior computational efficiency compared with their generative counterparts~\cite{DBLP:conf/iclr/ClarkLLM20}, it can be especially appealing to prompt-tuning discriminative PLMs. 

\begin{figure*}[t]
    \centering
    \includegraphics[width=\textwidth]{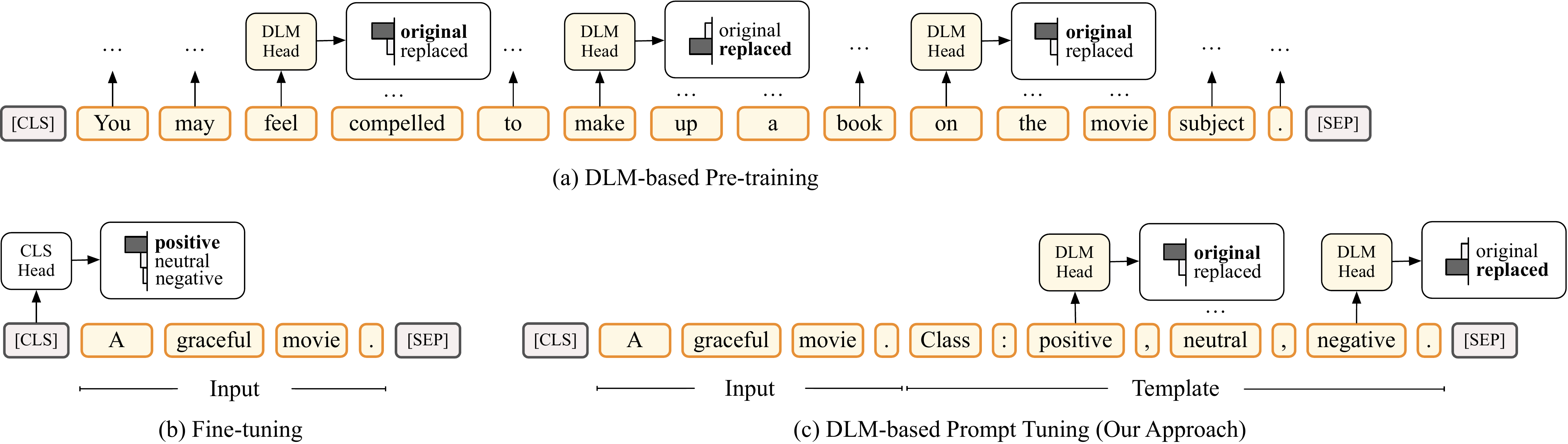}
    \caption{Illustration of (a) discriminative language modeling (DLM) based pre-training with the DLM head, (b)~vanilla fine-tuning with a new classification (CLS) head, and (c) our DPT prompt tuning approach that reformulates NLP tasks into a discriminative language modeling problem. DPT fills the input text into the template containing answer candidates, and discriminates whether each answer candidate is correct (i.e., original), or incorrect (i.e., replaced) based on the reused DLM head.}
    \vspace{-0.2em}
    \label{fig:framework}
\end{figure*}

In this work, we present DPT, the first \underline{p}rompt \underline{t}uning framework for \underline{d}iscriminative PLMs. DPT reformulates downstream tasks into a discriminative language modeling problem, maximally mitigating the gap between model pre-training and tuning. Specifically, as shown in Figure~\ref{fig:framework}, models are asked to discriminate correct answer tokens (e.g., correct labels for text classification, or answer spans for question answering) from the input tokens based on the reused discriminative language modeling head, where the objective form is identical to pre-training. 

To evaluate DPT, we conduct comprehensive experiments on text classification and question answering in both full-set and low-resource settings. Experimental results show that despite its simplicity, DPT significantly outperforms vanilla fine-tuning (e.g., $4.1\%$ accuracy improvement in the low-resource SST-5 evaluation). Moreover, previous works have shown that fine-tuning large PLMs can be highly unstable and even produce divergent results~\cite{DBLP:conf/naacl/DevlinCLT19,dodge2020fine}, which undermines the practicality of large PLMs. We show that DPT also addresses the unstable problem in tuning large discriminative PLMs.

The contributions of our work are summarized as follows: (1) We present the first prompt tuning framework for discriminative PLMs. (2) Comprehensive experimental results on text classification and question answering demonstrate the effectiveness of the proposed prompt tuning framework.

\section{Preliminary}
In this work, without loss of generality, we take ELECTRA~\cite{DBLP:conf/iclr/ClarkLLM20} as a representative example of discriminative PLMs, while applying DPT to other discriminative PLMs is also applicable. Here we introduce the main procedure of pre-training and fine-tuning, and we refer readers to the paper~\cite{DBLP:conf/iclr/ClarkLLM20} for more details.

\smallskip
\noindent
\textbf{Pre-training.} During pre-training, a generator first corrupts the text via token replacement. Then the discriminator is asked to detect the replaced tokens, by classifying each token into binary categories, i.e., $\{\rm{original}, \rm{replaced}\}$, as shown in Figure~\ref{fig:framework}. Finally, the generator is discarded and the discriminator is fine-tuned on downstream tasks. 


\smallskip
\noindent
\textbf{Vanilla Fine-tuning.} (1) During fine-tuning, to perform \textit{text classification}, a new classification head is typically introduced to classify the hidden representation of the \texttt{[CLS]} token in the last layer~\cite{DBLP:conf/iclr/ClarkLLM20}. (2) For general \textit{multi-span question answering}, the answer could be multiple spans from the input text~\cite{DBLP:conf/emnlp/DasigiLMSG19,DBLP:conf/naacl/DuaWDSS019}. State-of-the-art fine-tuning approaches formulate the task as a  sequence-labeling problem, and classify each input token into binary labels based on a new classification head, indicating whether the token belongs to the answer or not~\cite{DBLP:conf/emnlp/SegalESGB20,DBLP:conf/emnlp/YeLDLLSL20}. 

Note that the classification head typically introduces new parameters, and learning the parameters from scratch usually requires a large amount of labeled data. Moreover, previous works have shown that fine-tuning large PLMs can be highly unstable, and even produce divergent results~\cite{DBLP:conf/naacl/DevlinCLT19,dodge2020fine}. As a result, multiple fine-tuning trials are usually needed to find a good random seed that leads to a stably fine-tuned PLM, which undermines the practicality of large PLMs.

\section{Methodology}

In this section, we introduce the framework of DPT for prompt-tuning discriminative PLMs. We first introduce DPT using text classification as the running example, and then illustrate its application in question answering.

\smallskip
\noindent
\textbf{DLM-based Reformulation.} DPT reformulates NLP tasks into a dscriminative language modeling problem, maximally mitigating the gap between pre-training and tuning. Specifically, as shown in Figure~\ref{fig:framework}\hspace{0.5mm}(c), for a text classification task with class set $\mathcal{C}=\{c_1, c_2, \dots, c_n\}$, DPT defines a template that contains all answer candidates $\mathcal{T}(\cdot; \mathcal{C})$. Given an input text $x$ (e.g., ``\textit{A graceful movie.}''), DPT fills the input text into the template as follows:

{\fontsize{9.5pt}{9.5pt}\selectfont
\begin{equation}
    \mathcal{T}(x; \mathcal{C}) = \texttt{[CLS]} x \text{ Class: } c_1, c_2, \dots, c_n. \texttt{[SEP]} 
\end{equation}}

Intuitively, $\mathcal{T}(x; \mathcal{C})$ can be understood as creating a virtual context that assumes all candidate classes are correct for the input text $x$. It is then straightforward for discriminative PLMs to decide whether each class candidate token is proper in the context, by classifying the tokens into original (i.e., correct), or replaced (i.e., incorrect) based on the reused DLM head. In our experiments, we find that the order of classes in template has minimal influence on the performance, and a random order can produce good prompt-tuning results.

\smallskip
\noindent
\textbf{DPT Training.} After template filling, $\mathcal{T}(x; \mathcal{C})$ is fed into PLMs to obtain the hidden representations $\{\mathbf{h}_{\texttt{[CLS]}}, \mathbf{h}_1, \mathbf{h}_2, \dots, \mathbf{h}_m, \mathbf{h}_{\texttt{[SEP]}}\}$. PLMs are then prompted to discriminate whether each class is correct. Specifically, we compute the score of class $c_i$ based on the representation of the corresponding token $t_i$ as:\footnote{If the class name consists of multiple tokens, the representation of the first token is used.}

{\fontsize{10pt}{10pt}\selectfont
\begin{equation}
\label{eq:score}
s(c_i) = 1- \sigma(\mathbf{h}_\text{DLM}^\top \mathbf{h}_{t_i}),
\end{equation}}
\hspace{-1.1mm}where $\mathbf{h}_\text{DLM}$ is the reused DLM head, and $\sigma(\cdot)$ is the sigmoid activation. Note that in Equation~\ref{eq:score}, the computation of class scores is different from the vanilla fine-tuning approaches which encourage large inner products between the correct answer and classification head~\cite{DBLP:conf/naacl/DevlinCLT19,DBLP:conf/iclr/ClarkLLM20}. The rationale is that during pre-training, discriminative PLMs are typically required to produce large inner products for the replaced tokens (i.e., incorrect ones), and small inner products for the original tokens (i.e., correct ones)~\cite{DBLP:conf/iclr/ClarkLLM20}, and therefore Equation~\ref{eq:score} better fits the semantics in pre-training. In our experiments, we find this simple operation can lead to significantly better results in prompt-tuning discriminative PLMs. After obtaining the class score, the model is optimized as: 

{\fontsize{10pt}{10pt}\selectfont
\begin{equation}
    \mathcal{L} = \sum_{i}[ - y_i \log s(c_i) - (1 - y_i) \log (1 - s(c_i))],
\end{equation}}
\hspace{-1.4mm}where $y_i \in \{0, 1\}$ indicates the ground-truth label. Since DPT tunes PLMs by reusing the pre-trained DLM head in the same objective form as pre-training, compared with vanilla fine-tuning, we expect DPT will lead to more sample efficient and stable tuning results.

\smallskip
\noindent
\textbf{DPT for Question Answering.} Besides text classification, DPT can also be applied for the question answering task. Given a question and a paragraph, directly concatenating them without additional templates can already create a good prompting context. Then similar to text classification, we ask PLMs to discriminate whether each token in the paragraph is part of the answer (i.e., original), or not (i.e., replaced) based on the reused DLM head. During inference, we threshold the token scores to obtain multiple answer spans.

\section{Experiments}
In this section, we empirically evaluate DPT on the task of text classification and question answering.

\definecolor{olive}{RGB}{60,126,53}
\begin{table*}
    \begin{center}
    \small
    \begin{tabular}{lll|cccc|cccc}
    \toprule
    & \multirow{2}{*}{\hspace{-0.5em}PLM} & Tuning & \multicolumn{4}{c|}{Full-set Setting}& \multicolumn{4}{c}{Low-resource Setting} \\
    & & Approach & SST-2 & SST-5 & TREC & AGNews & SST-2 & SST-5 & TREC & AGNews  \\
    \midrule
    \parbox[t]{2mm}{\multirow{5}{*}{\rotatebox[origin=c]{90}{Base}}} 
    
    & \hspace{-0.5em}BERT & FT & 91.32 &  53.41 & 95.93 & 93.68 & 86.91 & 42.46 & 86.73 & 90.23\\
    & \hspace{-0.5em}RoBERTa & FT& 94.69 & 56.09 & 95.27 & 93.92 & 91.23 & 50.41 & 91.07 & 90.25\\
    & \hspace{-0.5em}ELECTRA & FT& 94.38 &  56.60 & 94.87 & 93.70 & 91.68 & 49.40 & 88.40 & 89.17\\
    & \hspace{-0.5em}ELECTRA & DPT (Ours) & \textbf{95.26} &  \textbf{58.34} & \textbf{96.27} & \textbf{94.22} & \textbf{93.83} & \textbf{53.48} & \textbf{93.93} & \textbf{90.60}\\
    &  & $\Delta$ & \hspace{-0.5mm}+\hspace{0.3mm}\textbf{0.88} &  \hspace{-0.5mm}+\hspace{0.3mm}\textbf{1.74} &  \hspace{-0.5mm}+\hspace{0.3mm}\textbf{1.40} & \hspace{-0.5mm}+\hspace{0.3mm}\textbf{0.52} & \hspace{-0.5mm}+\hspace{0.3mm}\textbf{2.15} & \hspace{-0.5mm}+\hspace{0.3mm}\textbf{4.08} & 
    \hspace{-0.5mm}+\hspace{0.3mm}\textbf{5.53} & \hspace{-0.5mm}+\hspace{0.3mm}\textbf{1.43}\\
    \midrule
    \parbox[t]{2mm}{\multirow{5}{*}{\rotatebox[origin=c]{90}{Large}}} 
    & \hspace{-0.5em}BERT& FT& 93.32 &  54.10 & 96.73 & 94.89 & 90.77 & 50.89 & 94.73 & 92.93\\
    & \hspace{-0.5em}RoBERTa & FT& 95.46 &  56.80 & 96.80 & 95.26 & 94.27 & 51.41 & 95.20 & 93.41\\
    & \hspace{-0.5em}ELECTRA & FT& 95.72 &  58.27 & 97.13 & 94.80 & 93.74 & 53.65 & 94.00 & 92.33\\
    & \hspace{-0.5em}ELECTRA & DPT (Ours)& \textbf{96.58} &  \textbf{60.69} & \textbf{98.07} & \textbf{95.38} & \textbf{96.09} & \textbf{57.00} & \textbf{95.67} & \textbf{93.58}\\
    &  & $\Delta$ & \hspace{-0.5mm}+\hspace{0.3mm}\textbf{0.86} &  \hspace{-0.5mm}+\hspace{0.3mm}\textbf{2.42} &  \hspace{-0.5mm}+\hspace{0.3mm}\textbf{0.94} & \hspace{-0.5mm}+\hspace{0.3mm}\textbf{0.58} & \hspace{-0.5mm}+\hspace{0.3mm}\textbf{2.35} & \hspace{-0.5mm}+\hspace{0.3mm}\textbf{3.35} & 
    \hspace{-0.5mm}+\hspace{0.3mm}\textbf{1.67} & \hspace{-0.5mm}+\hspace{0.3mm}\textbf{1.25}\\
    \bottomrule
    \end{tabular}
    \end{center}
    \caption{Experimental results on text classification. Full-set setting: 100$\%$ data, Low-resource setting: 10$\%$ data. FT: fine-tuning, DPT: discriminative prompt tuning. $\Delta$: Improvements of DPT over fine-tuning ELECTRA.}
    \label{table:main results}
\end{table*}

\begin{table}[t]
    \begin{center}
    \small
    \resizebox{\linewidth}{!}{%
    \begin{tabular}{ll|cc|cc}
    \toprule
    \multirow{2}{*}{PLM} & Tuning & \multicolumn{2}{c|}{Full Set}& \multicolumn{2}{c}{Low Resource} \\
    & Approach & EM & F1 & EM & F1 \\
    \midrule
    
    BERT & FT & 75.67 &  79.99 & 53.02 & 61.36 \\
    RoBERTa & FT & 78.29 &  84.56 & 59.31 & 67.56 \\
    ELECTRA & FT & 77.79 &  83.72 & 54.29 & 63.71 \\
    ELECTRA & DPT (Ours) & \textbf{79.66} &  \textbf{86.03} & \textbf{63.65} & \textbf{73.09} \\
    & $\Delta$ & \hspace{-0.5mm}+\hspace{0.3mm}\textbf{1.87} &  \hspace{-0.5mm}+\hspace{0.3mm}\textbf{2.31} & \hspace{-0.5mm}+\hspace{0.3mm}\textbf{9.36} & \hspace{-0.5mm}+\hspace{0.3mm}\textbf{9.38} \\
    \bottomrule
    \end{tabular}}
    \end{center}
    \caption{Experimental results of $\text{ELECTRA}_\text{large}$ on QUOREF multi-span question answering dataset.}
    \label{table:QA results}
\end{table}

\begin{table}[t]
\begin{center}
\resizebox{\linewidth}{!}{%
\begin{tabular}{lcccc}
\toprule
Tuning Approach & SST-2 &  SST-5 & TREC & AGNews \\
\midrule
Fine-tuning & 91.68 & 49.40 & 88.40 & 89.17\\
DPT ($\sigma$) & 92.16 & 50.96 & 88.00 & 90.29\\
DPT ($1-\sigma$) & \textbf{93.83} & \textbf{53.48} & \textbf{93.93} & \textbf{90.60}\\
\bottomrule
\end{tabular}}
\end{center}
\caption{Ablation on reuse forms of DLM head based on $\text{ELECTRA}_\text{base}$ in low-resource setting.}
\label{tab:ablation}
\end{table}

\smallskip
\noindent
\textbf{Datasets.} We evaluate DPT on four widely used text classification datasets, including SST-2, SST-5, TREC and AGNews. For question answering, we adopt the challenging QUOREF dataset, where for each question, there may exist multiple answer spans in the paragraph. We refer readers to Section~\ref{sec:dataset} for more dataset details.

\smallskip
\noindent
\textbf{Evaluation Protocols.} We evaluate the models under two settings, including (1) \textit{full-set} setting, where the full training data is available, and (2)~\textit{low-resource} setting, where only $10\%$ of the full training data for each dataset is available. We report the accuracy for text classification, and exact match (EM) and F1 score for question answering. To account for the unstable problem of baseline models, we report the average results from 3 best random seeds among 10 trials.

\smallskip
\noindent
\textbf{Baselines.} We compare DPT with several strong baseline models, including vanilla fine-tuning of ELECTRA~\cite{DBLP:conf/iclr/ClarkLLM20}, BERT~\cite{DBLP:conf/naacl/DevlinCLT19} and RoBERTa~\cite{DBLP:journals/corr/abs-1907-11692}. The fine-tuning of ELECTRA adopts the identical discriminative PLM to our model, and serves as the most direct baseline for comparison. 

\smallskip
\noindent
\textbf{Main Results.} We report the main results in Table~\ref{table:main results} and Table~\ref{table:QA results}, from which we observe that: (1) DPT significantly improves the performance of discriminative PLMs. The improvements are consistent across different tasks and datasets, as well as base and large models. (2) Previous works show that despite the significant improvements in low-resource setting, template-based prompt tuning typically can only approach fine-tuning performance in full-set setting~\cite{DBLP:conf/acl/GaoFC20}. In comparison, we note that DPT can improve the performance in both low-resource and full-set settings. The reason is that DPT enables PLMs to jointly model the input text and class candidates for better text understanding. In summary, DPT is effective in improving the performance of discriminative PLM tuning.

\begin{figure}[t]
\centering
\subfigure[Full-set Setting (100$\%$ data)]{
\label{fig:Fig1}
\includegraphics[width=0.98\linewidth]{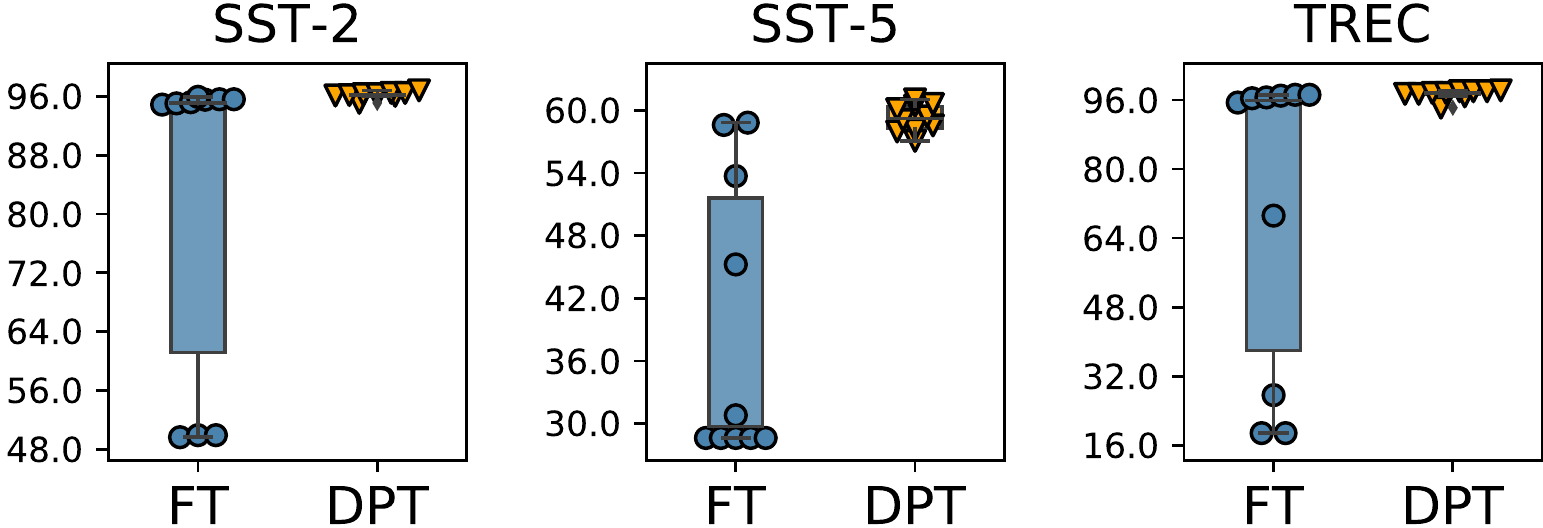}}
\subfigure[Low-resource Setting (10$\%$ data)]{
\label{fig:Fig2}
\includegraphics[width=0.98\linewidth]{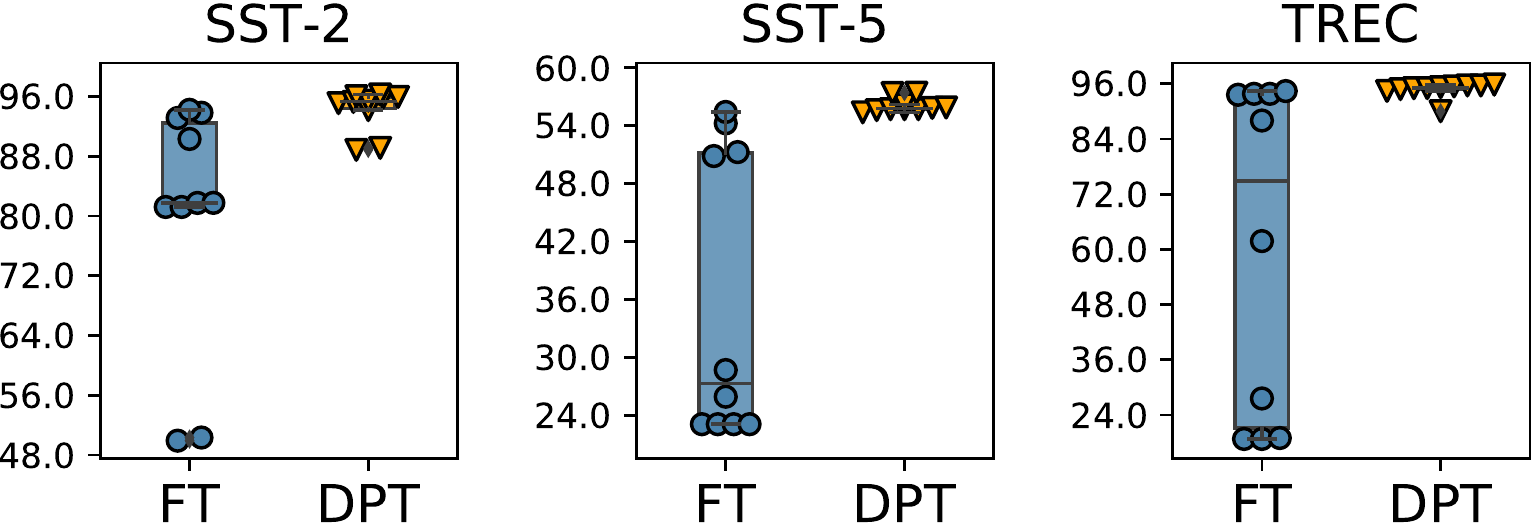}}
\caption{Performance distribution of $\text{ELECTRA}_\text{large}$ using fine-tuning and DPT from 10 seeds.}
\label{fig:stable}
\end{figure}

\smallskip
\noindent
\textbf{Tuning Stability.} Previous works have commonly observed the instability of fine-tuning large generative PLMs~\cite{DBLP:conf/naacl/DevlinCLT19,dodge2020fine}. Some works attempt to alleviate the problem by careful initialization and optimization~\cite{DBLP:conf/iclr/0007WKWA21}, or intermediate fine-tuning on other large-scale datasets~\cite{DBLP:journals/corr/abs-1811-01088}. To investigate the tuning stability of discriminative PLMs, we tune $\text{ELECTRA}_\text{large}$ using fine-tuning and DPT from 10 random seeds. From the results in Figure~\ref{fig:stable}, we observe that: (1)~Similar to generative PLMs, fine-tuning large discriminative PLMs is also highly unstable, and can even frequently produce divergent results (e.g., nearly $20\%$ accuracy for 5-way classification in SST-5 in low-resource setting). The problem is exacerbated by sparse data in low-resource setting, but remains even in full-set setting. (2) DPT achieves significantly more stable tuning results in both full-set and low-resource settings, where all tuning trials converged and closely approach the best performance. This is due to the reuse of DLM head parameters and identical objective forms to pre-training. 

\smallskip
\noindent
\textbf{Ablation Study.} In DPT, different from conventional fine-tuning approaches, correct labels are encouraged to have small inner products with classifiers (as indicated by the $1-\sigma$ in Equation~\ref{eq:score}). We evaluate DPT using conventional score computation (i.e., $\sigma$), and report the results in Table~\ref{tab:ablation}. The significant drop in performance shows that a proper form of reusing DLM head is crucial to the results of prompt-tuning discriminative PLMs.



\section{Conclusion and Future Work}
In this work, we present a simple and effective prompt tuning approach for discriminative PLMs. We note directly performing large-scale classification (e.g., for hundreds of classes) with DPT may be computationally inefficient. In future, we plan to address the problem by classifying text following class hierarchies, where each hierarchical layer typically consists of a moderate number of classes.

\section{Acknowledgement}
This work is suooprted by the Natural Science Foundation of China (NSFC) and the German Research Foundation (DFG) in Project Crossmodal Learning, NSFC 61621136008 / DFC TRR-169, Institute Guo Qiang at Tsinghua University, and International Innovation Center of Tsinghua University, Shanghai, China.

\bibliography{acl}

\begin{thebibliography}{25}
\expandafter\ifx\csname natexlab\endcsname\relax\def\natexlab#1{#1}\fi

\bibitem[{Brown et~al.(2020)Brown, Mann, Ryder, Subbiah, Kaplan, Dhariwal,
  Neelakantan, Shyam, Sastry, Askell et~al.}]{brown2020language}
Tom~B Brown, Benjamin Mann, Nick Ryder, Melanie Subbiah, Jared Kaplan, Prafulla
  Dhariwal, Arvind Neelakantan, Pranav Shyam, Girish Sastry, Amanda Askell,
  et~al. 2020.
\newblock \href {https://arxiv.org/abs/2005.14165} {Language models are
  few-shot learners}.
\newblock \emph{arXiv preprint arXiv:2005.14165}.

\bibitem[{Clark et~al.(2020)Clark, Luong, Le, and
  Manning}]{DBLP:conf/iclr/ClarkLLM20}
Kevin Clark, Minh{-}Thang Luong, Quoc~V. Le, and Christopher~D. Manning. 2020.
\newblock \href {https://openreview.net/forum?id=r1xMH1BtvB} {{ELECTRA:}
  pre-training text encoders as discriminators rather than generators}.
\newblock In \emph{Proceedings of ICLR}.

\bibitem[{Dasigi et~al.(2019)Dasigi, Liu, Marasovic, Smith, and
  Gardner}]{DBLP:conf/emnlp/DasigiLMSG19}
Pradeep Dasigi, Nelson~F. Liu, Ana Marasovic, Noah~A. Smith, and Matt Gardner.
  2019.
\newblock \href {https://doi.org/10.18653/v1/D19-1606} {{QUOREF}: {A} reading
  comprehension dataset with questions requiring coreferential reasoning}.
\newblock In \emph{Proceedings of EMNLP-IJCNLP}, pages 5924--5931.

\bibitem[{Devlin et~al.(2019)Devlin, Chang, Lee, and
  Toutanova}]{DBLP:conf/naacl/DevlinCLT19}
Jacob Devlin, Ming{-}Wei Chang, Kenton Lee, and Kristina Toutanova. 2019.
\newblock \href {https://doi.org/10.18653/v1/n19-1423} {{BERT:} pre-training of
  deep bidirectional transformers for language understanding}.
\newblock In \emph{Proceedings of NAACL-HLT}, pages 4171--4186.

\bibitem[{Dodge et~al.(2020)Dodge, Ilharco, Schwartz, Farhadi, Hajishirzi, and
  Smith}]{dodge2020fine}
Jesse Dodge, Gabriel Ilharco, Roy Schwartz, Ali Farhadi, Hannaneh Hajishirzi,
  and Noah Smith. 2020.
\newblock \href {https://arxiv.org/abs/2002.06305} {Fine-tuning pretrained
  language models: Weight initializations, data orders, and early stopping}.
\newblock \emph{arXiv preprint arXiv:2002.06305}.

\bibitem[{Dua et~al.(2019)Dua, Wang, Dasigi, Stanovsky, Singh, and
  Gardner}]{DBLP:conf/naacl/DuaWDSS019}
Dheeru Dua, Yizhong Wang, Pradeep Dasigi, Gabriel Stanovsky, Sameer Singh, and
  Matt Gardner. 2019.
\newblock \href {https://doi.org/10.18653/v1/n19-1246} {{DROP:} {A} reading
  comprehension benchmark requiring discrete reasoning over paragraphs}.
\newblock In \emph{Proceedings of NAACL-HLT}, pages 2368--2378.

\bibitem[{Gao et~al.(2021)Gao, Fisch, and Chen}]{DBLP:conf/acl/GaoFC20}
Tianyu Gao, Adam Fisch, and Danqi Chen. 2021.
\newblock \href {https://doi.org/10.18653/v1/2021.acl-long.295} {Making
  pre-trained language models better few-shot learners}.
\newblock In \emph{Proceedings of ACL-IJCNLP}, pages 3816--3830.

\bibitem[{Lan et~al.(2020)Lan, Chen, Goodman, Gimpel, Sharma, and
  Soricut}]{DBLP:conf/iclr/LanCGGSS20}
Zhenzhong Lan, Mingda Chen, Sebastian Goodman, Kevin Gimpel, Piyush Sharma, and
  Radu Soricut. 2020.
\newblock \href {https://openreview.net/forum?id=H1eA7AEtvS} {{ALBERT:} {A}
  lite {BERT} for self-supervised learning of language representations}.
\newblock In \emph{Proceedings of ICLR}.

\bibitem[{Liu et~al.(2021)Liu, Yuan, Fu, Jiang, Hayashi, and
  Neubig}]{DBLP:journals/corr/abs-2107-13586}
Pengfei Liu, Weizhe Yuan, Jinlan Fu, Zhengbao Jiang, Hiroaki Hayashi, and
  Graham Neubig. 2021.
\newblock \href {https://arxiv.org/abs/2107.13586} {Pre-train, prompt, and
  predict: A systematic survey of prompting methods in natural language
  processing}.
\newblock \emph{arXiv preprint arXiv:2107.13586}.

\bibitem[{Liu et~al.(2019)Liu, Ott, Goyal, Du, Joshi, Chen, Levy, Lewis,
  Zettlemoyer, and Stoyanov}]{DBLP:journals/corr/abs-1907-11692}
Yinhan Liu, Myle Ott, Naman Goyal, Jingfei Du, Mandar Joshi, Danqi Chen, Omer
  Levy, Mike Lewis, Luke Zettlemoyer, and Veselin Stoyanov. 2019.
\newblock \href {http://arxiv.org/abs/1907.11692} {Roberta: A robustly
  optimized bert pretraining approach}.
\newblock \emph{arXiv preprint arXiv:1907.11692}.

\bibitem[{Petroni et~al.(2019)Petroni, Rockt{\"{a}}schel, Riedel, Lewis,
  Bakhtin, Wu, and Miller}]{DBLP:conf/emnlp/PetroniRRLBWM19}
Fabio Petroni, Tim Rockt{\"{a}}schel, Sebastian Riedel, Patrick S.~H. Lewis,
  Anton Bakhtin, Yuxiang Wu, and Alexander~H. Miller. 2019.
\newblock \href {https://doi.org/10.18653/v1/D19-1250} {Language models as
  knowledge bases?}
\newblock In \emph{Proceedings of EMNLP-IJCNLP}, pages 2463--2473.

\bibitem[{Phang et~al.(2018)Phang, F{\'e}vry, and
  Bowman}]{DBLP:journals/corr/abs-1811-01088}
Jason Phang, Thibault F{\'e}vry, and Samuel~R Bowman. 2018.
\newblock \href {http://arxiv.org/abs/1811.01088} {Sentence encoders on
  {STILTS}: Supplementary training on intermediate labeled-data tasks}.
\newblock \emph{arXiv preprint arXiv:1811.01088}.

\bibitem[{Qiu et~al.(2020)Qiu, Sun, Xu, Shao, Dai, and Huang}]{qiu2020pre}
Xipeng Qiu, Tianxiang Sun, Yige Xu, Yunfan Shao, Ning Dai, and Xuanjing Huang.
  2020.
\newblock \href {https://arxiv.org/abs/2003.08271} {Pre-trained models for
  natural language processing: A survey}.
\newblock \emph{SCTS}, pages 1--26.

\bibitem[{Raffel et~al.(2020)Raffel, Shazeer, Roberts, Lee, Narang, Matena,
  Zhou, Li, and Liu}]{DBLP:journals/jmlr/RaffelSRLNMZLL20}
Colin Raffel, Noam Shazeer, Adam Roberts, Katherine Lee, Sharan Narang, Michael
  Matena, Yanqi Zhou, Wei Li, and Peter~J. Liu. 2020.
\newblock \href {http://jmlr.org/papers/v21/20-074.html} {Exploring the limits
  of transfer learning with a unified text-to-text transformer}.
\newblock \emph{JMLR}, 21:140:1--140:67.

\bibitem[{Schick and Sch{\"{u}}tze(2021)}]{DBLP:conf/naacl/SchickS21}
Timo Schick and Hinrich Sch{\"{u}}tze. 2021.
\newblock \href {https://doi.org/10.18653/v1/2021.naacl-main.185} {It's not
  just size that matters: Small language models are also few-shot learners}.
\newblock In \emph{Proceedings of NAACL-HLT}, pages 2339--2352.

\bibitem[{Segal et~al.(2020)Segal, Efrat, Shoham, Globerson, and
  Berant}]{DBLP:conf/emnlp/SegalESGB20}
Elad Segal, Avia Efrat, Mor Shoham, Amir Globerson, and Jonathan Berant. 2020.
\newblock \href {https://doi.org/10.18653/v1/2020.emnlp-main.248} {A simple and
  effective model for answering multi-span questions}.
\newblock In \emph{Proceedings of EMNLP}, pages 3074--3080.

\bibitem[{Socher et~al.(2013)Socher, Perelygin, Wu, Chuang, Manning, Ng, and
  Potts}]{socher-etal-2013-recursive}
Richard Socher, Alex Perelygin, Jean Wu, Jason Chuang, Christopher~D. Manning,
  Andrew Ng, and Christopher Potts. 2013.
\newblock \href {https://www.aclweb.org/anthology/D13-1170} {Recursive deep
  models for semantic compositionality over a sentiment treebank}.
\newblock In \emph{Proceedings of EMNLP}, page 1631–1642.

\bibitem[{Voorhees and Tice(2000)}]{DBLP:conf/sigir/VoorheesT00}
Ellen~M. Voorhees and Dawn~M. Tice. 2000.
\newblock \href {https://doi.org/10.1145/345508.345577} {Building a question
  answering test collection}.
\newblock In \emph{Proceedings of SIGIR}, pages 200--207.

\bibitem[{Xiong et~al.(2020)Xiong, Du, Wang, and
  Stoyanov}]{DBLP:conf/iclr/XiongDWS20}
Wenhan Xiong, Jingfei Du, William~Yang Wang, and Veselin Stoyanov. 2020.
\newblock \href {https://openreview.net/forum?id=BJlzm64tDH} {Pretrained
  encyclopedia: Weakly supervised knowledge-pretrained language model}.
\newblock In \emph{Proceedings of ICLR}.

\bibitem[{Xu et~al.(2021)Xu, Zhengyan et~al.}]{xu2021pre}
Han Xu, Zhang Zhengyan, et~al. 2021.
\newblock \href {https://arxiv.org/pdf/2106.07139.pdf} {Pre-trained models:
  Past, present and future}.
\newblock \emph{arXiv preprint arXiv:2106.07139}.

\bibitem[{Yang et~al.(2019)Yang, Dai, Yang, Carbonell, Salakhutdinov, and
  Le}]{DBLP:conf/nips/YangDYCSL19}
Zhilin Yang, Zihang Dai, Yiming Yang, Jaime~G. Carbonell, Ruslan Salakhutdinov,
  and Quoc~V. Le. 2019.
\newblock \href
  {https://proceedings.neurips.cc/paper/2019/hash/dc6a7e655d7e5840e66733e9ee67cc69-Abstract.html}
  {{XLNet}: Generalized autoregressive pretraining for language understanding}.
\newblock In \emph{Proceedings of NeurIPS}, pages 5754--5764.

\bibitem[{Yao et~al.(2021)Yao, Zhang, Zhang, Liu, Chua, and Sun}]{yao2021cpt}
Yuan Yao, Ao~Zhang, Zhengyan Zhang, Zhiyuan Liu, Tat-Seng Chua, and Maosong
  Sun. 2021.
\newblock {CPT}: Colorful prompt tuning for pre-trained vision-language models.
\newblock \emph{arXiv preprint arXiv:2109.11797}.

\bibitem[{Ye et~al.(2020)Ye, Lin, Du, Liu, Li, Sun, and
  Liu}]{DBLP:conf/emnlp/YeLDLLSL20}
Deming Ye, Yankai Lin, Jiaju Du, Zhenghao Liu, Peng Li, Maosong Sun, and
  Zhiyuan Liu. 2020.
\newblock \href {https://doi.org/10.18653/v1/2020.emnlp-main.582}
  {Coreferential reasoning learning for language representation}.
\newblock In \emph{Proceedings of EMNLP}, pages 7170--7186.

\bibitem[{Zhang et~al.(2021)Zhang, Wu, Katiyar, Weinberger, and
  Artzi}]{DBLP:conf/iclr/0007WKWA21}
Tianyi Zhang, Felix Wu, Arzoo Katiyar, Kilian~Q. Weinberger, and Yoav Artzi.
  2021.
\newblock \href {https://openreview.net/forum?id=cO1IH43yUF} {Revisiting
  few-sample {BERT} fine-tuning}.
\newblock In \emph{Proceedings of ICLR}.

\bibitem[{Zhang et~al.(2015)Zhang, Zhao, and LeCun}]{NIPS2015_250cf8b5}
Xiang Zhang, Junbo Zhao, and Yann LeCun. 2015.
\newblock \href
  {https://proceedings.neurips.cc/paper/2015/file/250cf8b51c773f3f8dc8b4be867a9a02-Paper.pdf}
  {Character-level convolutional networks for text classification}.
\newblock In \emph{Proceedings of NIPS}, pages 649--657.

\end{thebibliography}
\bibliographystyle{acl_natbib}

\appendix

\section{Implementation Details}
In this work, we take ELECTRA~\cite{DBLP:conf/iclr/ClarkLLM20} as an representative example of discriminative PLMs, including (1) $\text{ELECTRA}_\text{base}$ with $768$ dimensional hidden representations, $12$ encoding layers and $110$M parameters, and (2) $\text{ELECTRA}_\text{large}$ with $1,024$ dimensional hidden representations, $24$ encoding layers and $340$M parameters. 

For text classification tasks, we follow the hyperparameters in \citet{DBLP:conf/iclr/ClarkLLM20}, and train the base models for $10$ epochs with learning rate $2$e-$5$ and batchsize $32$ on $2$ GeForce RTX 2080 Ti GPUs. And we train the large models for $10$ epochs with learning rate $2$e-$5$ and batchsize $8$ on $2$ GeForce RTX 2080 Ti GPUs. For question answering, we follow the hyparameters in \citet{DBLP:conf/emnlp/SegalESGB20}, and train the large models for $20$ epochs with learning rate $5$e-$6$ and batchsize $2$ on $6$ GeForce RTX 2080 Ti GPUs. During inference, a token is considered as part of the answer if its score is lower than $0.6$.

\section{Dataset Details}
\label{sec:dataset}
We evaluate DPT on four popular text classification datasets, including SST-2~\cite{socher-etal-2013-recursive}, SST-5~\cite{socher-etal-2013-recursive}, TREC~\cite{DBLP:conf/sigir/VoorheesT00} and AGNews~\cite{NIPS2015_250cf8b5}. For question answering task, we adopt the challenging QUOREF dataset~\cite{DBLP:conf/emnlp/DasigiLMSG19}, where there may exist multiple answers in the paragraph for each question. Specifically, QUOREF contains $21,817$ questions and $4,225$ paragraphs, where each question has $1.15$ answers on average. The average length for the questions and paragraphs are $15.49$ and $325.68$ respectively. We report the results on the validation set for QUOREF, since its test set is not publicly available, and report the results on the test set for the other datasets.

\section{Further Results of Tuning Stability}
\begin{figure}[t]
\centering
\subfigure[Full Set. ]{\includegraphics[scale=0.5]{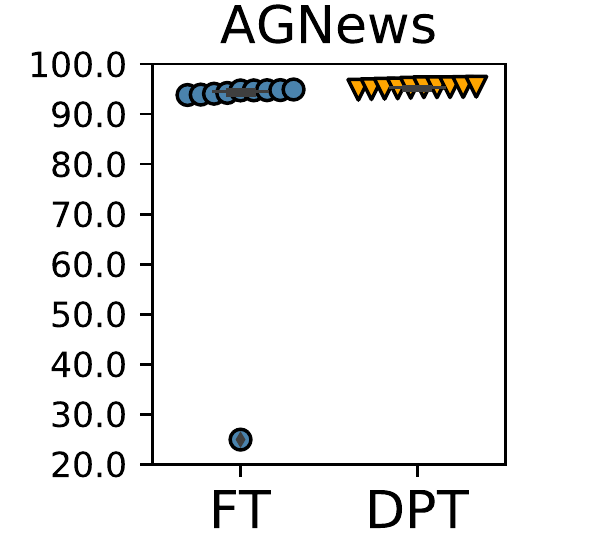}}
\subfigure[Low Resource.]{\includegraphics[scale=0.5]{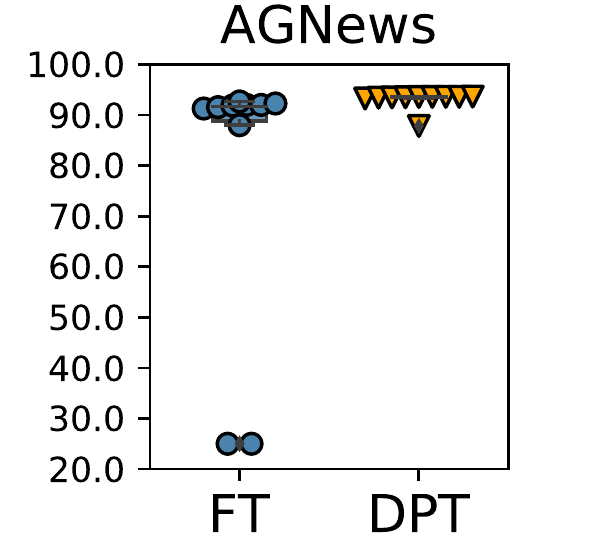}}

\caption{Performance distribution of $\text{ELECTRA}_\text{large}$ using fine-tuning and DPT from 10 seeds.}
\label{fig:AGNEWS}

\end{figure}

We report the performance distribution of AGNews in Figure~\ref{fig:AGNEWS}. We observe that the unstable problem of fine-tuning large discriminative PLMs remains even for the large-scale AGNews dataset with 120K training samples. The results show the advantage of DPT in stably tuning discriminative PLMs.

\end{document}